%% file: root.tex
\documentclass[letterpaper, 10 pt, conference]{ieeeconf}  

\usepackage{blindtext}
\usepackage{graphicx}
\graphicspath{ {images/} }

\usepackage[export]{adjustbox}
\usepackage{cite}
\usepackage{amssymb}  
\usepackage{calc}
\usepackage[strings]{underscore}
\usepackage{cleveref}
\usepackage[cmex10]{amsmath}
\usepackage{array}
\newcolumntype{L}[1]{>{\raggedright\let\newline\\\arraybackslash\hspace{0pt}}m{#1}}
\newcolumntype{C}[1]{>{\centering\let\newline\\\arraybackslash\hspace{0pt}}m{#1}}
\newcolumntype{R}[1]{>{\raggedleft\let\newline\\\arraybackslash\hspace{0pt}}m{#1}}
\usepackage{mathtools}
\usepackage[font=footnotesize,subrefformat=parens,labelformat=parens]{subfig}
\usepackage{algorithm}
\usepackage[noend]{algpseudocode}
\usepackage{setspace}
\makeatletter
\def\BState{\State\hskip-\ALG@thistlm}
\makeatother

\IEEEoverridecommandlockouts                              

\title{\LARGE \bf
UAV Pose Estimation using Cross-view Geolocalization \\ with Satellite Imagery
}

\author{Akshay Shetty and Grace Xingxin Gao
\thanks{Both authors are with the Aerospace Engineering Department, University of Illinois at Urbana-Champaign, IL 61820, USA {\tt\small \{ashetty2, gracegao\}@illinois.edu}}
}

\begin{document}

\maketitle
\thispagestyle{empty}
\pagestyle{empty}

\setlength{\belowcaptionskip}{-4pt}

\begin{abstract}
\input{abstract.tex}
\end{abstract}

\section{INTRODUCTION}
\input{introduction.tex}

\section{DATASET}
\input{dataset.tex}

\section{APPROACH}
\input{approach.tex}

\section{RESULTS}
\input{results.tex}

\section{CONCLUSIONS}
\input{conclusion.tex}


\section*{ACKNOWLEDGMENT}
The authors would like to thank Google for publicly sharing their Google Maps and Google Earth imagery.


\bibliographystyle{IEEEtran}
\bibliography{thesisrefs}

\end{document}

%% file: abstract.tex
We propose an image-based cross-view geolocalization method that estimates the global pose of a UAV with the aid of georeferenced satellite imagery. Our method consists of two Siamese neural networks that extract relevant features despite large differences in viewpoints. The input to our method is an aerial UAV image and nearby satellite images, and the output is the weighted global pose estimate of the UAV camera. We also present a framework to integrate our cross-view geolocalization output with visual odometry through a Kalman filter. We build a dataset of simulated UAV images and satellite imagery to train and test our networks. We show that our method performs better than previous camera pose estimation methods, and we demonstrate our networks ability to generalize well to test datasets with unseen images. Finally, we show that integrating our method with visual odometry significantly reduces trajectory estimation errors. 


%% file: introduction.tex
Global pose estimation is essential for autonomous outdoor navigation of unmanned aerial vehicles (UAVs). Currently, Global Positioning System (GPS) is primarily relied on for outdoor positioning. However, in certain cases GPS signals might be erroneous or unavailable due to reflections or blockages from nearby structures, or due to malicious spoofing or jamming attacks. In such cases additional on-board sensors such as cameras are desired to aid navigation. Visual odometry has been extensively studied to estimate camera pose from a sequence of images \cite{mur2017orb, 6907054, 6906584, 6096039}. However, in the absence of global corrections, odometry accumulates drift which can be significant for long trajectories. Loop closure methods have been proposed to mitigate the drift \cite{6224843, 6386145, 5980273}, but they require the camera to visit a previous scene from the trajectory. Another method to reduce visual odometry drift is to perform image-based geolocalization.

In recent years, deep learning has been widely used for image-based geolocalization tasks and has been shown to out-perform traditional image matching and retrieval methods. These methods can be categorized into either single-view geolocalization \cite{weyand2016planet,vo2017revisiting,kendall2015posenet} or cross-view geolocalization \cite{lin2013cross,workman2015wide,lin2015learning,vo2016localizing,kim2017learned,tian2017cross,kim2017satellite}. Single-view geolocalization implies that the method geolocates a query image using georeferenced images from a similar view. PlaNet \cite{weyand2016planet} and the revisited IM2GPS \cite{vo2017revisiting} are two methods that apply deep learning to obtain coarse geolocation up to a few kilometers. PoseNet \cite{kendall2015posenet} estimates camera pose on a relatively local scale within a region previously covered by the camera. However, it might not always be feasible to have a georeferenced database of images from the same viewpoint as the query images.

The other category of cross-view geolocalization implies that the method geolocates a query image using georeferenced images from a different view. These methods have been used to geolocate ground-based query images using a database of satellite imagery. Lin et al. \cite{lin2013cross} train kernels to geolocalize a ground image using satellite images and land cover attribute maps. Workman et al. \cite{workman2015wide} train convolutional neural networks (CNN) to extract meaningful, multi-scale features from satellite images to match with the corresponding ground image features. Lin et al. \cite{lin2015learning} train a Siamese network \cite{bromley1994signature} with contrastive loss \cite{hadsell2006dimensionality} to match Google street-view images to oblique satellite imagery. Vo et al. \cite{vo2016localizing} train a similar Siamese network while presenting methods to increase the rotational invariance of the matching process. Kim et al. \cite{kim2017learned} improve on past geolocalization results by training a reweighting network to highlight important query image features. Tian et al. \cite{tian2017cross} first detect buildings in the query and reference images, followed by using a trained Siamese network and dominant sets to find the best match. Kim et al. \cite{kim2017satellite} combine the output of a Siamese network with a particle filter to estimate the georeferenced vehicle pose from a sequence of input images.

\begin{figure}[t!]
\includegraphics[width=\linewidth]{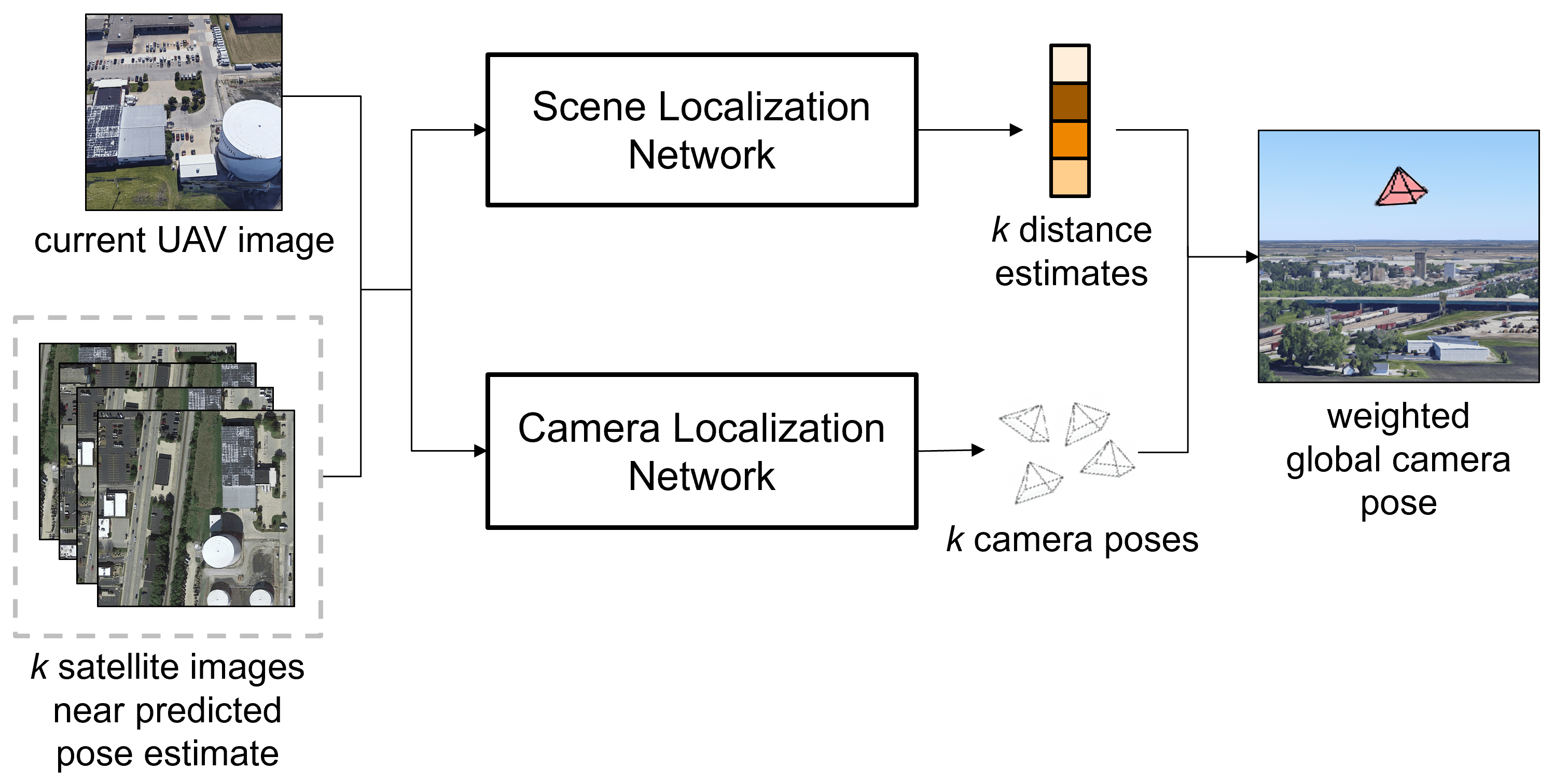}
\centering
\caption{Our cross-view geolocalization method combines outputs from scene and camera localization networks to obtain a global camera pose estimate.}
\label{fig:cross-view-geolocalization}
\end{figure}

We approach the problem of estimating the UAVs global pose from its aerial image as a cross-view geolocalization problem. The above cross-view geolocalization methods focus on localizing the scene depicted in the image and not the camera. Previous methods to estimate the camera pose include certain assumptions, such as assuming a fixed forward distance from the query image center to the reference image \cite{kim2017satellite}. While such an assumption seems practical for a ground vehicle, it is not realistic for UAV-based images which could be captured at varying tilt (pitch) angles. Additionally, UAV global pose estimation has more degrees of freedom compared to ground vehicle pose estimation. This paper investigates the performance of deep learning for global camera pose estimation.

\begin{figure}[t!]
    \centering
  \subfloat[]{%
       \includegraphics[width=\linewidth]{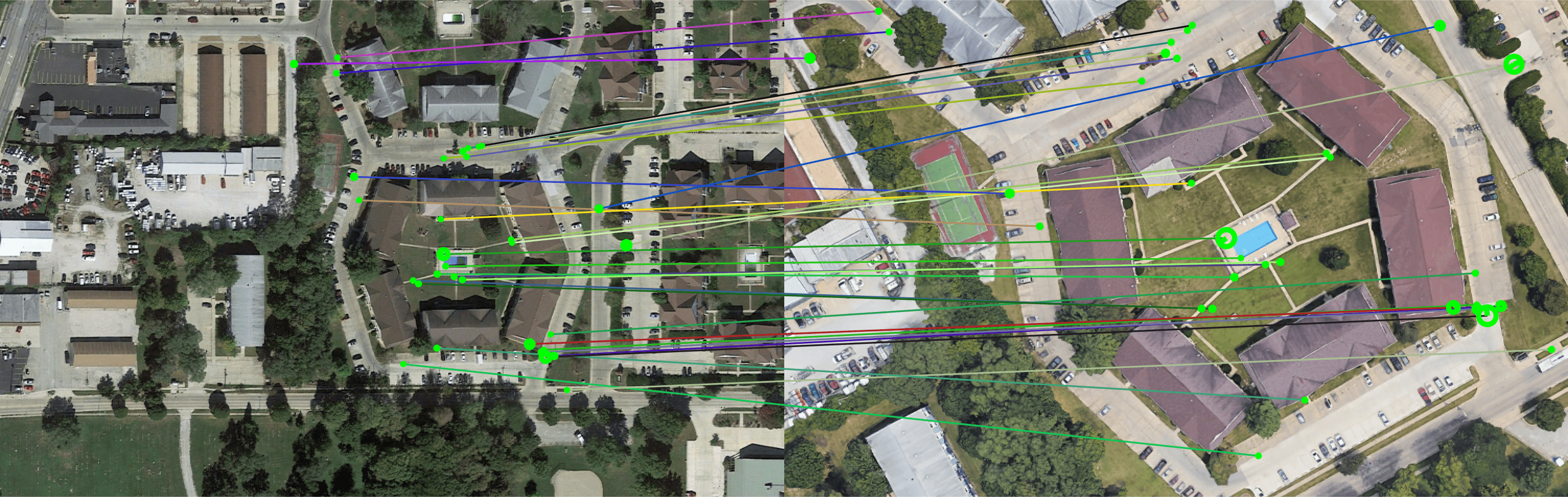}}
    \label{fig:sift-working}\hfill
  \\
  \subfloat[]{%
        \includegraphics[width=\linewidth]{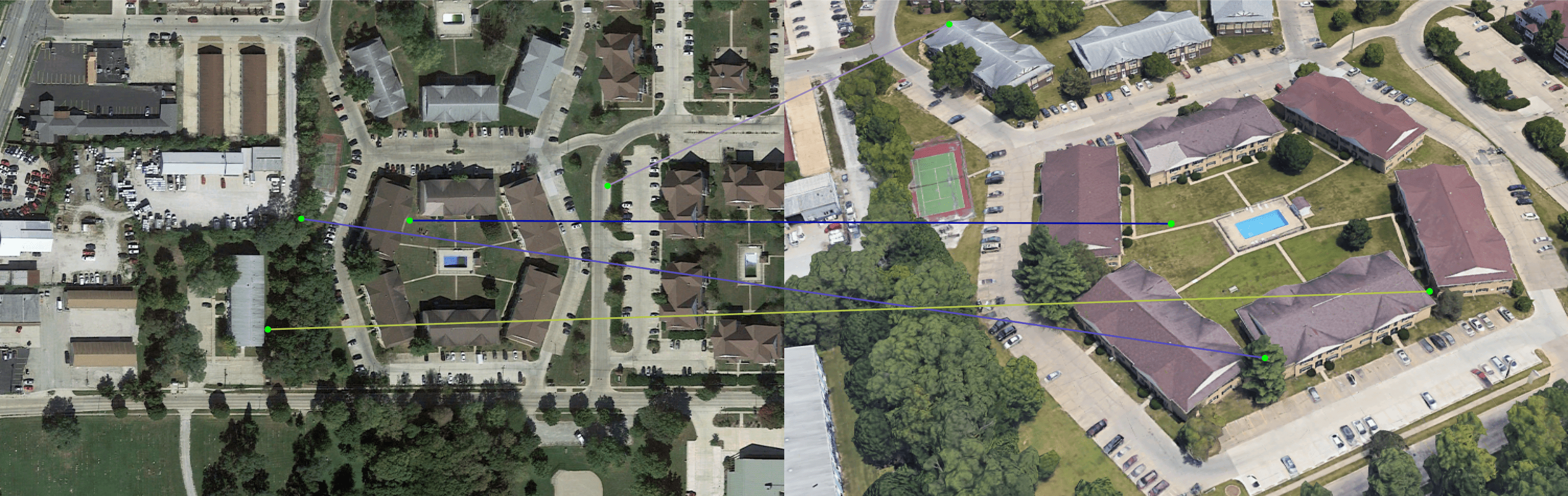}}
    \label{fig:sift-not-working}
  \caption{Traditional feature extraction methods work well for short baseline images (a), but fail for large changes in viewpoint (b).}
  \label{fig:sift-viewpoint}
\end{figure}

In this paper, we propose a cross-view geolocalization method to estimate the UAVs global pose with the aid of georeferenced satellite imagery as shown in Figure \ref{fig:cross-view-geolocalization}. Based on the recent progress of the community, we use deep learning in order to extract relevant features from cross-view images. Figure \ref{fig:sift-viewpoint} shows that traditional feature extraction algorithms struggle for our cross-view dataset. Our cross-view geolocalization method consists of two networks: a scene localization network and a camera localization network. We combine the output from both these networks to obtain a weighted global camera pose estimate. We also provide a Kalman filter (KF) framework for integrating our cross-view geolocalization network with visual odometry. We compare the performance of our method with the regression network proposed by Melekhov et al. \cite{melekhov2017relative}. We evaluate our method on a dataset of UAV and satellite images built by us.

%% file: dataset.tex
\begin{figure}[b!]
\includegraphics[width=\linewidth]{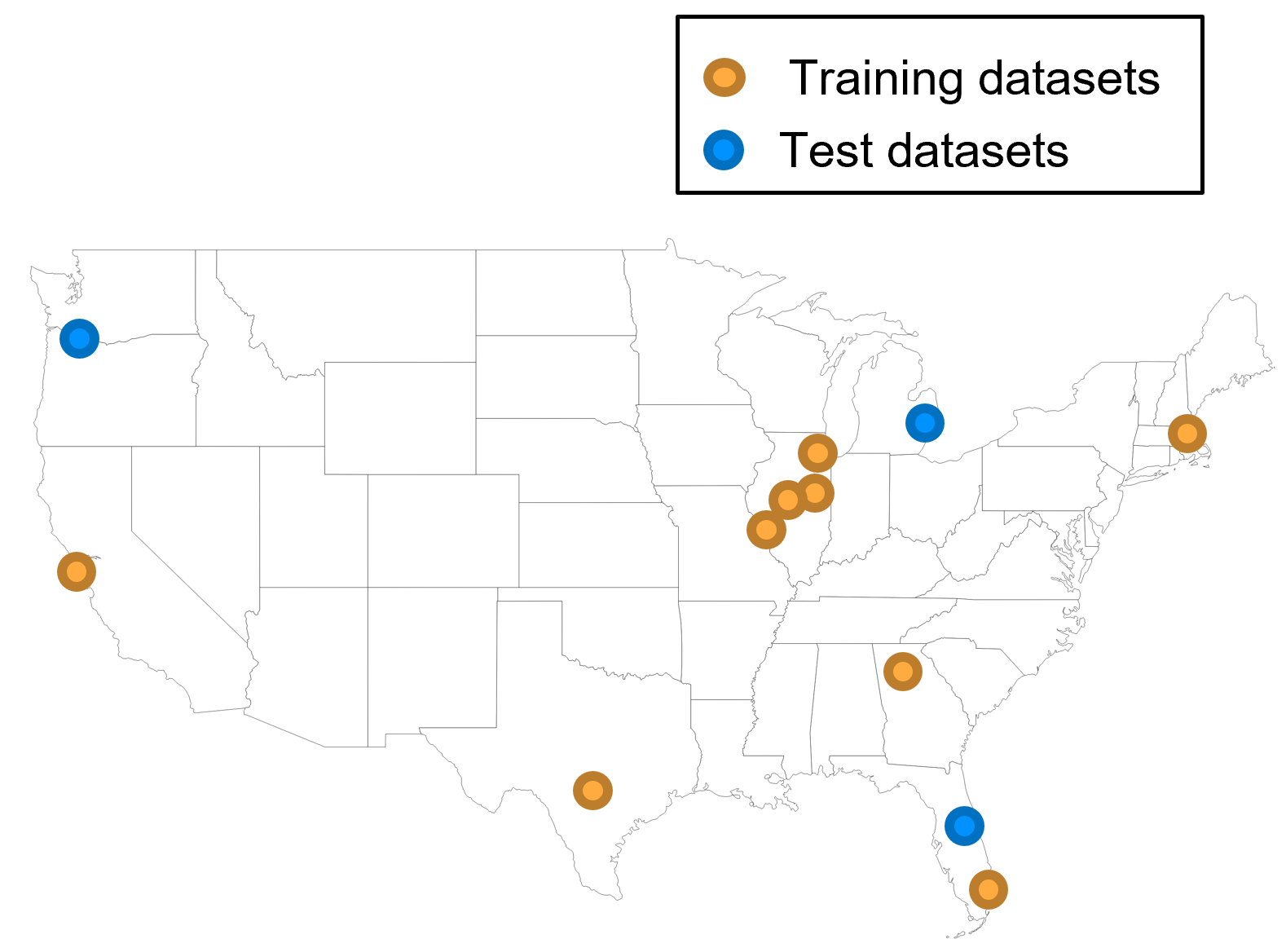}
\centering
\caption{We generate UAV and satellite image pairs from 12 different cities across US. 9 of these datasets are used for training our networks and the remaining 3 are used for testing our networks}
\label{fig:dataset-map}
\end{figure}
\begin{figure}[t!]
    \centering
  \subfloat[]{%
       \includegraphics[width=\linewidth]{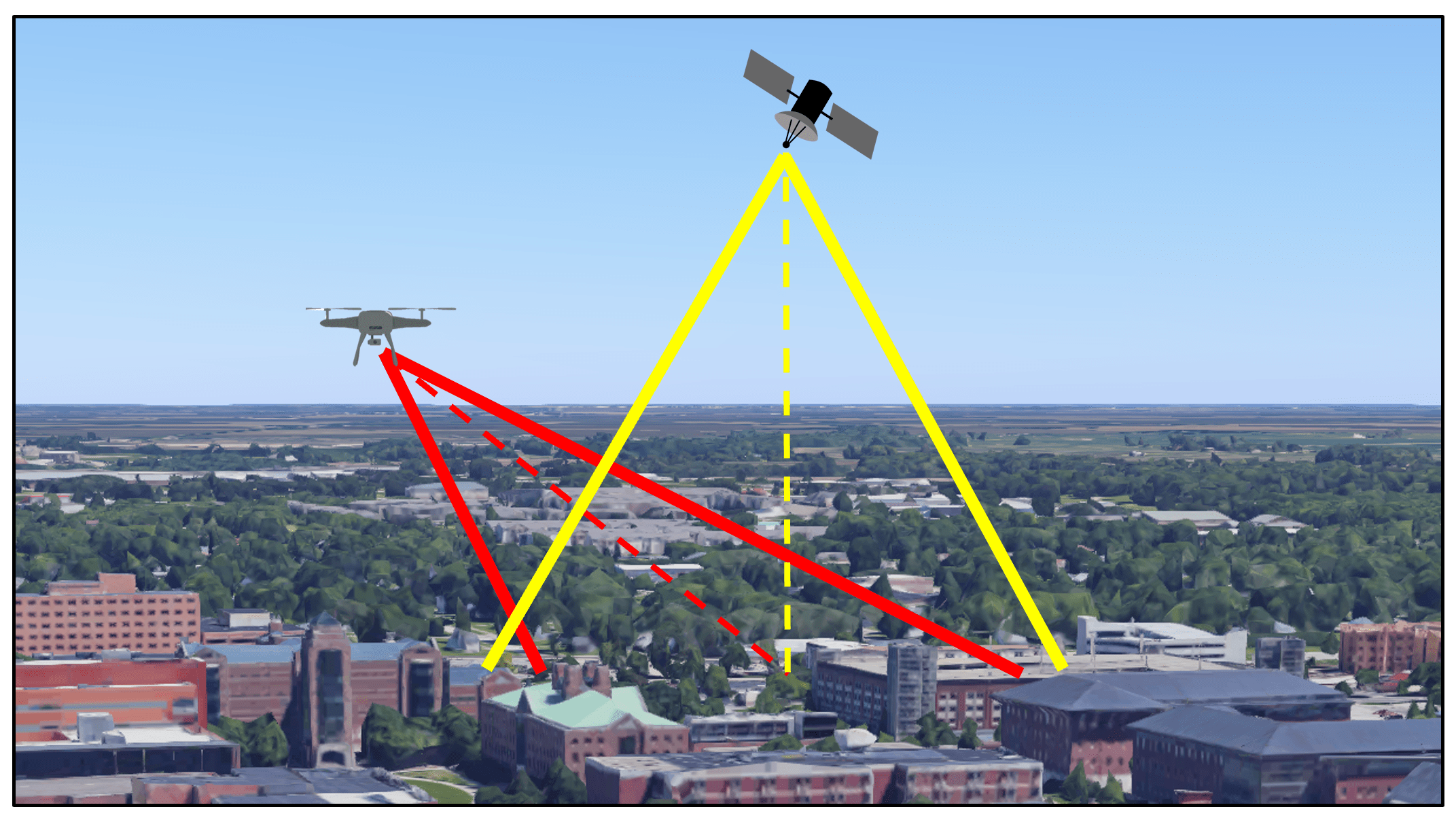}}
  \\
  \subfloat[]{%
        \includegraphics[width=\linewidth]{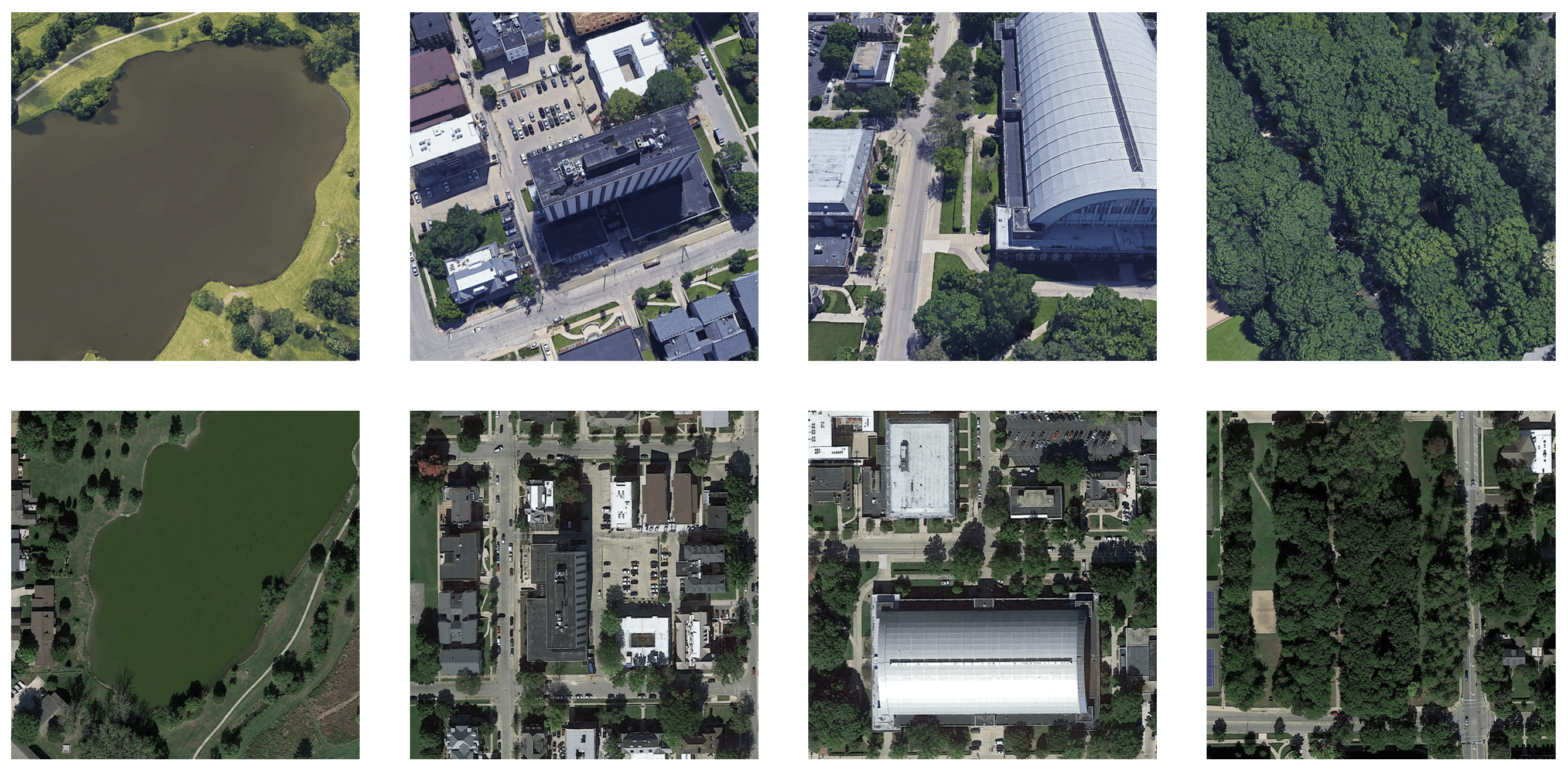}}
  \caption{UAV and satellite image pairs. (a) Relationship between our UAV and satellite images. (b) Sample matching pairs of UAV images from Google Earth (top row) and satellite images from Google Maps (bottom row).}
  \label{fig:dataset-pairs}
\end{figure}

For our dataset, we create pairs of UAV and satellite images. We extract our satellite images from Google Maps, and use Google Earth to simulate our UAV images. The images from both the datasets are created from different distributions \cite{ge_gm_diff} as shown in Figure \ref{fig:dataset-pairs}(b). In order to have a good distribution of the images and to increase the generalization capability of our networks, we build our dataset from 12 different cities across US as shown in Figure \ref{fig:dataset-map}. For our training and validation datasets, we use image pairs from Atlanta, Austin, Boston, Champaign (IL), Chicago, Miami, San Francisco, Springfield (IL) and St. Louis. For our test dataset, we use image pairs from Detroit, Orlando and Portland.

We first create the UAV images, followed by the satellite images. For each of the above cities, we define a cuboid in space within which we sample uniformly random 3D positions to generate the UAV images. The altitude of the cuboid ranges from $100m$ to $200m$ above ground, in order to prevent any sample position from colliding with Google Earth objects. For a downward facing camera, the heading (yaw) and the roll axes are aligned. Thus, we set the roll angle to $0^0$ in order to prevent any ambiguity in orientation. Additionally, we limit the tilt (pitch) of the camera to between $0^0$ and $45^0$ to ensure that the image sufficiently captures the ground scenery. Thus, each UAV image is defined by $\{ x, y, z, \psi, \theta \}$, where $x, y,$ and $z$ represents the 3D position within the corresponding cuboid, $\psi$ is the heading $\in [-180^0, 180^0]$ and $\theta$ is the tilt $\in [0^0, 45^0]$. 

Next, for each of the UAV images we extract a corresponding satellite image from Google Maps. We set the altitude for all satellite images to $300m$ above ground. We set the horizontal position of the satellite image as the point where the UAV image center intersects the ground altitude. All the satellite images have $0^0$ heading, ie. are aligned with the North direction. Figure \ref{fig:dataset-pairs} shows the relationship between a UAV and satellite image pair, and shows some sample image pairs. The labels for each image pair contain the camera transformation between the satellite image and the UAV image, ie. the relative 3D position, the heading and tilt angles.

For the training and validation datasets we generate $\sim27K$ and $\sim3K$ image pairs respectively. For the test dataset we generate $\sim3K$ image pairs. To reduce overfitting in our networks, we augment our training dataset by rotating the satellite images by $90^0, 180^0$ and $270^0$. Thus, we finally obtain an augmented training dataset of $\sim108K$ image pairs.


In order to test the integration of our cross-view geolocalization with visual odometry, we simulate a flight trajectory in the city of Detroit. We generate $4000$ consecutive UAV images resulting in a trajectory length of $\sim2.5km$ and duration of $\sim200s$. The initial part of the trajectory was chosen to be pure translation in order to ensure good initialization of the visual odometry algorithm. We generate satellite images at every $50m$ for the flight region, resulting in $\sim1.7K$ satellite images.

%% file: approach.tex
Our cross-view geolocalization method consists of two separate networks: a scene localization network, and a camera localization network. The scene localization network estimates the distance between a UAV-satellite image pair, whereas the camera localization network estimates the camera transformation between image pair. Finally, we combine the output from the two networks to obtain a weighted global camera pose estimate as shown in Figure \ref{fig:cross-view-geolocalization}. We also provide a Kalman filter framework to integrate our cross-view geolocalization output with visual odometry.

\subsection{Scene Localization}

\begin{figure}[b!]
\includegraphics[width=\linewidth]{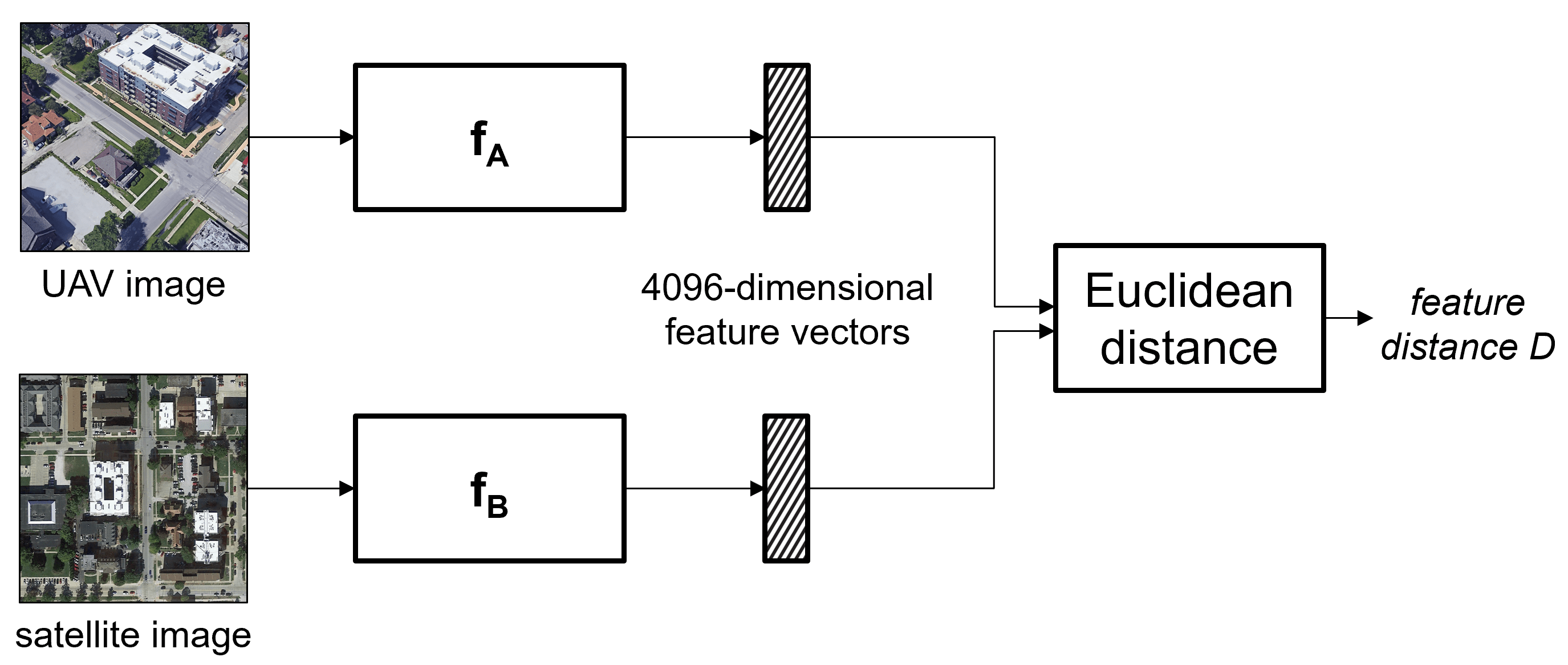}
\centering
\caption{Our scene localization network. We implement a Siamese architecture to extract 4096-dimensional feature vectors from the UAV and satellite images. We train the network with a contrastive loss function in order to map features closer from a matching pair.}
\label{fig:scene-loc-network}
\end{figure}

For our scene localization network, we use a Siamese architecture \cite{bromley1994signature} similar to previous methods \cite{lin2015learning,vo2016localizing,tian2017cross,kim2017satellite}. Each branch of our network, $\mathbf{f_A}$ and $\mathbf{f_B}$, consists of an AlexNet \cite{krizhevsky2012imagenet} architecture without the final classification layer. We use a different set of parameters for both the branches since the images come from different distributions. The output from each branch is a $4096$-dimensional feature vector as shown in Figure \ref{fig:scene-loc-network}. Finally, we calculate the euclidean distance between the two feature vectors to get the scene localization network output:

\begin{equation}
    d = \text{dist}( \mathbf{f_A}(I_U), \mathbf{f_B}(I_S) ),
\end{equation}
where $\mathbf{f_A}$ and $\mathbf{f_B}$ represent the operations that the AlexNet layers apply on the UAV image $I_U$ and the satellite image $I_S$. We train the network using the constrastive loss function \cite{hadsell2006dimensionality} as follows:

\begin{equation}
    L_{scene} = l\ast d^2 + (1-l)\ast (\max\{0, m - d\})^2 ,
\end{equation}
where $l$ is a pair indicator ($l=0$ for non-matching pair, and $l=1$ for matching pair) and $m$ is the feature distance margin. For a matching pair, the network learns to reduce the distance between the feature vectors. For a non-matching pair, the network learns to push the feature vectors further apart if their distance is less than $m$.

We initialize each branch of the network with weights and biases pretrained on the Places365 dataset \cite{zhou2017places}. For training the network we use the $108K$ image pairs in the training dataset. Additionally, for each UAV image we randomly select a non-matching satellite image which is at least $100m$ away horizontally. Thus, we finally have $108K$ matching image pairs and $108K$ non-matching image pairs for training the scene localization network. We set the feature distance margin $m$ to $100$ and use mini-batch stochastic gradient descent \cite{ruder2016overview} with a batch size of $128$ ($64$ matching and $64$ non-matching pairs). We begin with a learning rate of $10^{-5}$ decayed by a factor of $0.7$ every $4$ epochs.

\subsection{Camera Localization}

For our camera localization network, we again use a Siamese architecture with a hybrid classification and regression structure to estimate the camera transformation. Based on the method we used to create our dataset, the relative horizontal position coordinates lie in the range $x \in [-200m, 200m]$, $y \in [-200m, 200m]$. We divide the horizontal plane into 64 cells and classify the horizontal position into one of the cells. We regress the remaining camera transformation variables. Each branch of the Siamese architecture, $\mathbf{f_C}$ and $\mathbf{f_D}$, consists of only the convolutional layers of AlexNet. The output from each branch is $9216$-dimensional feature vector extracted by the convolutional layers. The feature vectors from the branches are concatenated and input to two sets of fully connected layers. The first set of fully connected layers outputs a $64-dimensional$ vector $\mathbf{c}_{out}$ that corresponds to the $64$ horizontal position cells. The cell coordinates corresponding to the maximum of $\mathbf{c}_{out}$ is used as the horizontal position estimate $(\widehat{x}, \widehat{y})$. The second set of fully connected layers regresses the vertical position, heading and tilt to obtain $\widehat{z}$, $\widehat{\psi}$ and $\widehat{\theta}$ respectively. Finally, we obtain the 3D position estimate as $\widehat{\mathbf{p}} = (\widehat{x}, \widehat{y}, \widehat{z})$.

\begin{figure}[b!]
\includegraphics[width=\linewidth]{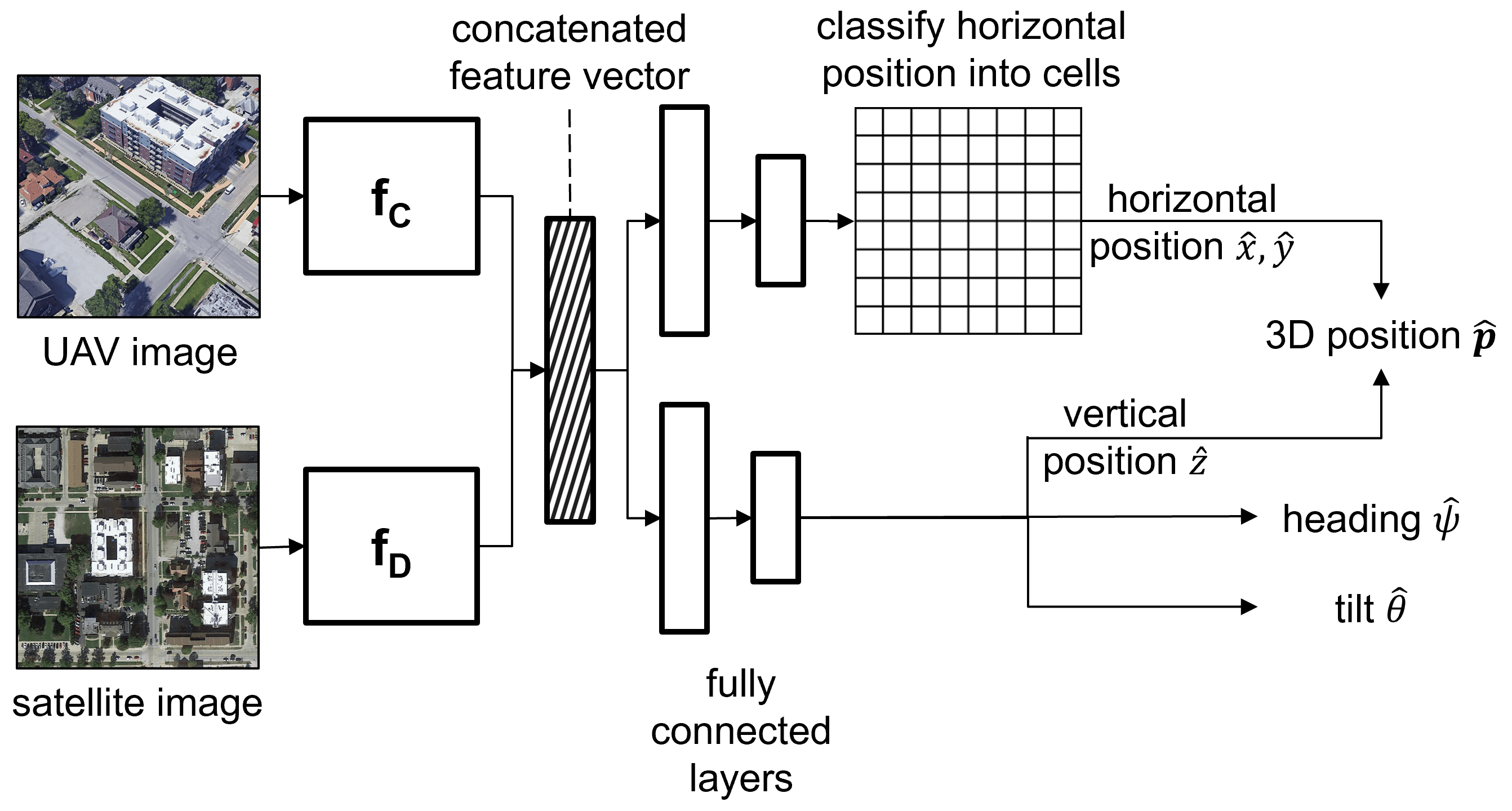}
\centering
\caption{Our camera localization network. We implement a Siamese architecture to estimate the UAV pose. The network classifies the horizontal position into 1 of 64 cells, and regresses the UAV vertical position and orientation.}
\label{fig:camera-loc-network}
\end{figure}

For training the network, we use the cross entropy loss for the horizontal position cell classification, and absolute errors for vertical position, heading and tilt:

\begin{equation}
\begin{split}
L_{xy} & = -\log\left(\frac{\exp(\mathbf{c}_{out}[c_l])}{\sum_{j=0}^{63} \exp(\mathbf{c}_{out}[j])}\right) \\
L_z & = | z_l - \widehat{z} | \\
L_{\psi} & = | \psi_l - \widehat{\psi} | \\
L_{\theta} & = | \theta_l - \widehat{\theta} |,
\end{split}
\end{equation}
where $c_l, z_l, \psi_l$ and $\theta_l$ represent the respective target values. We combine the individual loss functions to obtain the total loss function to train our camera localization network:

\begin{equation}
    L_{camera} = \alpha L_{xy} + L_z + \beta L_{\psi} + \gamma L_{\theta},
\end{equation}
where $\alpha, \beta$ and $\gamma$ are weighing parameters fixed during training.

Similar to the scene localization network, we initialize each branch of our network with weights and biases pretrained on the Places365 dataset \cite{zhou2017places}. For training the network we use the $108K$ image pairs in the training dataset. Initially we train only the classification section of the network to ensure convergence, followed by fine-tuning the complete network. We use mini-batch stochastic gradient descent \cite{ruder2016overview} with a batch size of $64$. For the weighing parameters, we test different values and find the following to perform best: $\alpha = 30$, $\beta = 1.0$ and $\gamma = 0.5$. We begin with a learning rate of $3 \times 10^{-5}$ decayed by a factor of $0.7$ every $4$ epochs.

\subsection{Cross-view Geolocalization}
To obtain the weighted global camera pose from our networks as shown in Figure \ref{fig:cross-view-geolocalization}, we implement the following steps:

\begin{itemize}
    \item For the current step, first obtain the predicted position estimate $\mathbf{p}_{t|t-1}$ from the KF. Choose $k$ nearest satellite images from the database: $I_S^1, I_S^2, \cdots , I_S^k$. Use the current UAV image $I_U$ to generate the following image pairs: $(I_U,I_S^1), (I_U,I_S^2), \cdots , (I_U,I_S^k)$
    
    \item Pass the $k$ image pairs through the scene localization network to obtain $k$ distance values: $d^1, d^2, \cdots, d^k$.
    
    \item Pass the $k$ image pairs through the camera localization network to obtain $k$ camera pose estimates: $(\widehat{\mathbf{p}}^1, \widehat{\psi}^1, \widehat{\theta}^1), (\widehat{\mathbf{p}}^2, \widehat{\psi}^2, \widehat{\theta}^2), \cdots , (\widehat{\mathbf{p}}^k, \widehat{\psi}^k, \widehat{\theta}^k)$.
    
    \item Based on the $k$ distance values and $k$ camera pose estimates, calculate a weighted global camera pose estimate as follows:
    
\begin{equation}
\label{eq:weighted-pose}
\hspace{-0.6cm} \bar{\mathbf{p}} = \frac{\sum_{i=1}^{k} \left(\frac{\widehat{\mathbf{p}}^i}{d_i}\right)}{\sum_{i=1}^{k} \left(\frac{1}{d_i}\right)}, \bar{\psi} = \frac{\sum_{i=1}^{k} \left(\frac{\widehat{\psi}^i}{d_i}\right)}{\sum_{i=1}^{k} \left(\frac{1}{d_i}\right)}, \bar{\theta} = \frac{\sum_{i=1}^{k} \left(\frac{\widehat{\theta}^i}{d_i}\right)}{\sum_{i=1}^{k} \left(\frac{1}{d_i}\right)}
\end{equation}
    
    \item Use the distribution of the $k$ camera pose estimates as the covariance global camera pose estimate to be used in KF correction step:
    
\begin{equation}
\label{eq:weighted-pose-cov}
\mathbf{M} =
\left[
\arraycolsep=1.4pt\def\arraystretch{1.5}
\begin{array}{c|c|c}
\text{cov}(\widehat{\mathbf{p}}^i) & 0 & 0 \\
\hline
0 & \text{cov}(\widehat{\psi}^i) & 0 \\
\hline
0 & 0 & \text{cov}(\widehat{\theta}^i)
\end{array}
\right]
\end{equation}
\end{itemize}

\subsection{Integrating with Visual Odometry}
\label{subsec:integrate-VO}
\begin{figure}[t!]
\includegraphics[width=\linewidth]{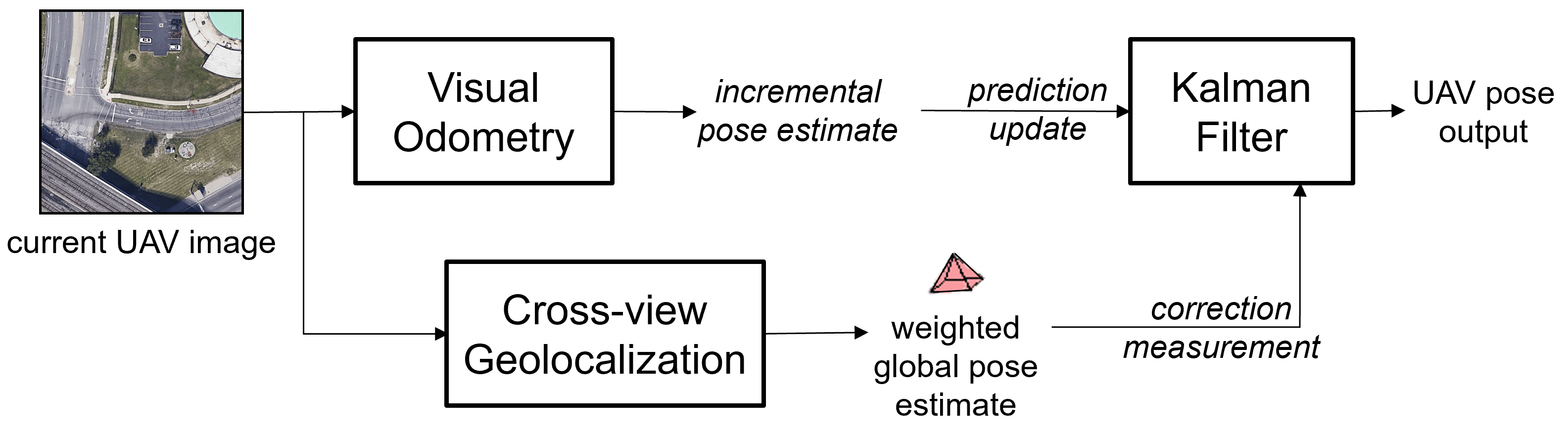}
\centering
\caption{Our approach for fusing the global pose output from our cross-view geolocalization with visual odometry via a Kalman filter.}
\label{fig:fusion-approach}
\end{figure}

In this section we explain our method for integrating our cross-view geolocalization with visual odometry. We use a KF where we use the visual odometry in the prediction step and the cross-view geolocalization output in the correction step as shown in Figure \ref{fig:fusion-approach}.

\begin{table*}
\centering
 \begin{tabular}{| c | C{2cm} | C{2cm} | C{2cm} | C{2cm} | C{2cm} |} 
 \hline
   & Horizontal error (m) & Vertical error (m) & 3D position error (m) & Heading error (deg) & Tilt error (deg) \\ [0.5ex] 
 \hline
 \multicolumn{6}{|l|}{Validation dataset:} \\ [0.5ex] 
 \hline
 Camera Regression \cite{melekhov2017relative} & 45.75 & 17.00 & 48.86 & 37.66 & 7.9 \\
 \hline
 \textbf{Camera Hybrid} & 33.23 & 15.72 & 36.81 & 32.07 & 6.08\\ 
 \hline
 \multicolumn{6}{|l|}{Test dataset:} \\ [0.5ex] 
 \hline
 Camera Regression \cite{melekhov2017relative} & 68.06 & 17.32 & 70.70 & 70.64 & 7.94 \\
 \hline
 \textbf{Camera Hybrid} & 33.86 & 16.05 & 37.55 & 31.68 & 6.28\\ 
 \hline
\end{tabular}
\caption{Camera localization results for the validation and test datasets. Our hybrid network performs better than the previous regression method and generalizes well to unseen test dataset.}
\label{tab:camera-loc-val-results}
\vspace{0.2cm}
\end{table*}

For visual odometry we implement the open-source ORB-SLAM2 \cite{mur2017orb} package on our image sequence. To obtain the calibration file required for ORB-SLAM2, we overlay the standard checkerboard \cite{bradski2008learning} on Google Earth and capture screenshots from different viewpoints. Additionally, we scale the ORB-SLAM2 output using the true position information for the first 100m of the trajectory. After scaling the ORB-SLAM2 trajectory output, we extract incremental position estimates $\Delta \mathbf{p}_{VO}$ and incremental rotation matrix $\Delta \mathbf{R}_{VO}$ between consecutive images.

Our state vector for the KF consists of the 6D UAV camera pose: $\mathbf{X} = \{ \mathbf{p}, \psi, \theta, \phi \}$. We represent the orientation as euler angles in order to easily perform the correction step using our cross-view geolocalization output. We set the process noise covariance matrix as $\mathbf{Q} = \text{diag}([0.01, 0.01, 0.01, 0.01, 0.01, 0.01])$. We use the incremental pose estimate from visual odometry in our prediction step:

\begin{equation}
\begin{split}
    \mathbf{p}_{t|t-1} & = \mathbf{p}_{t-1} + \Delta \mathbf{p}_{VO} \\
    \{ \psi, \theta, \phi \}_{t|t-1} & = T^e_R ( \Delta \mathbf{R}_{VO} \ast T^R_e(\{ \psi, \theta, \phi \}_{t-1}) ) \\
    \mathbf{P_{t|t-1}} & = \mathbf{P_{t-1|t-1}} + \mathbf{Q},
\end{split}
\end{equation}
where $T^e_R$ represents the transformation from rotation matrices to euler angles, and $T^R_e$ represents the transformation from euler angles to rotation matrices. $\mathbf{P}$ represents the state covariance matrix.

For the correction step, we use our weighted global camera pose output from equation (\ref{eq:weighted-pose}) : $\mathbf{z}_{t} = \{ \bar{\mathbf{p}}, \bar{\psi}, \bar{\theta} \}$, and its covariance $\mathbf{M}$ from equation (\ref{eq:weighted-pose-cov}). Thus the observation model turns out to be as follows:
\begin{equation}
    \mathbf{H} = \begin{bmatrix}
1 & 0 & 0 & 0 & 0 & 0\\ 
0 & 1 & 0 & 0 & 0 & 0\\ 
0 & 0 & 1 & 0 & 0 & 0\\ 
0 & 0 & 0 & 1 & 0 & 0\\ 
0 & 0 & 0 & 0 & 1 & 0
\end{bmatrix}
\end{equation}

The correction step is performed as follows:

\begin{equation}
\begin{split}
\mathbf{y}_t & = \mathbf{z}_t - \mathbf{H}\mathbf{X}_{t|t-1} \\
\mathbf{S}_t & = \mathbf{M}_t + \mathbf{H}\mathbf{P}_{t|t-1}\mathbf{H}^T \\
\mathbf{K}_t & = \mathbf{P}_{t|t-1}\mathbf{H}^T\mathbf{S}_t^{-1} \\
\mathbf{X}_{t|t} & = \mathbf{X}_{t|t-1} + \mathbf{K}_t\mathbf{y}_t \\
\mathbf{P}_{t|t} & = (\mathbf{I} - \mathbf{K}_t\mathbf{H})\mathbf{P}_{t|t-1}
\end{split}
\end{equation}

%% file: results.tex
\subsection{Scene and Camera Localization}

\begin{figure}
    \centering
    \includegraphics[width=\linewidth]{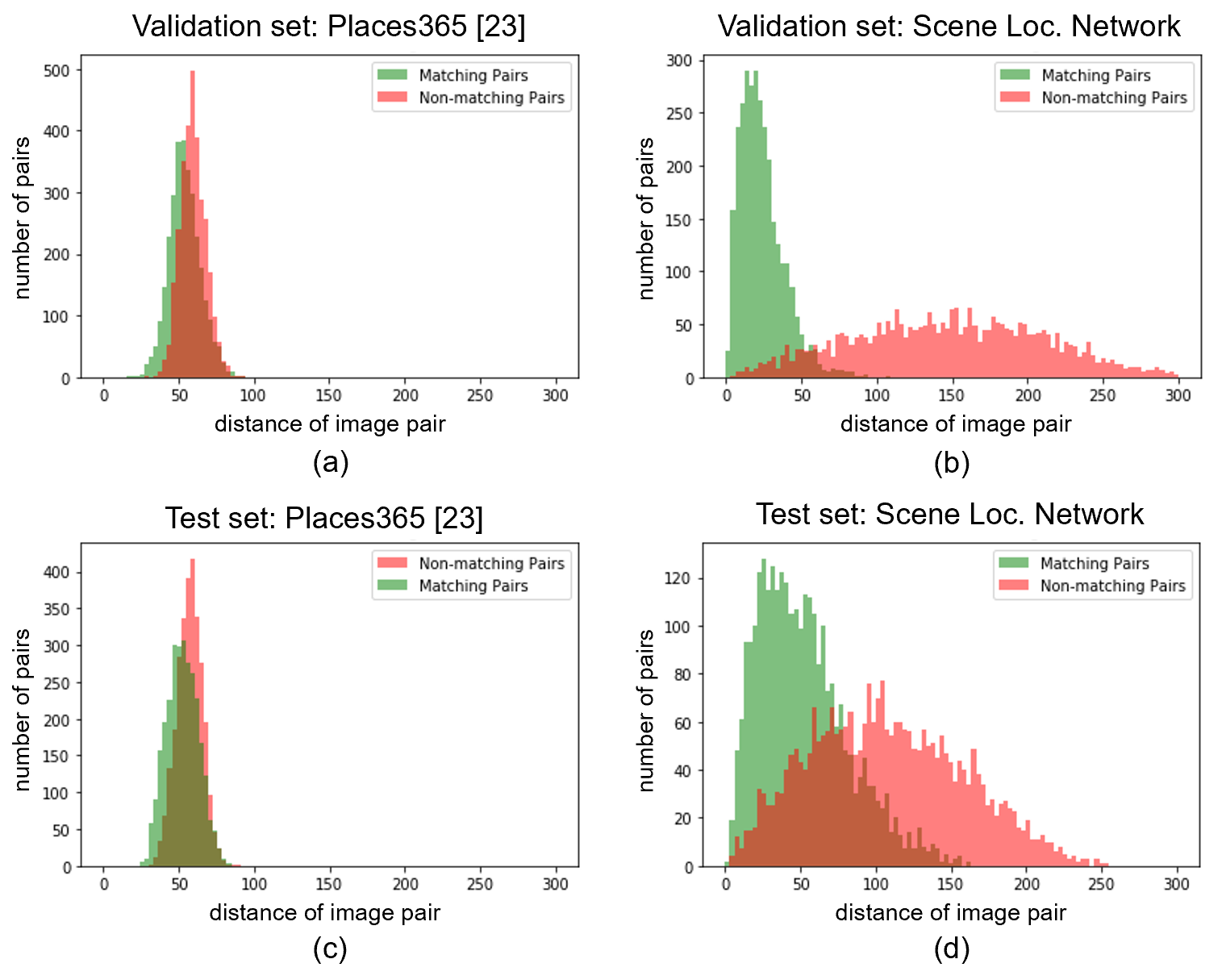}
    \caption{Comparison of feature distance histograms for the validation (a), (b) and test (c), (d) datasets. (a) and (c) are from the Places-365 network \cite{zhou2017places} not trained on the contrastive loss yet, and (b) and (d) are from our trained scene localization network.}
    \label{fig:scene-loc-results}
\end{figure}

In this section, we check the performance of our scene and camera localization networks on the validation and test datasets. The validation dataset is from the same distribution of images as the training dataset, whereas the test dataset comprises image pairs from new cities the networks have not seen before. Figure \ref{fig:scene-loc-results} shows the feature distance histograms obtained from the scene localization network and the Places-365 \cite{zhou2017places} network. We observe that, based on the contrastive loss training, our scene localization network learns to generate closer feature vectors for matching image-pairs and distant feature vectors for non-matching pairs. 


For our camera localization network, we use the previous work by Melekhov et al. \cite{melekhov2017relative} as a baseline for comparison. In \cite{melekhov2017relative} the authors use Siamese architecture followed by a single fully connected layer to regress the camera transformation between the two input images. The branches of the Siamese architecture consist of the AlexNet convolutional layers, except for the final pooling layer. For the remainder of the paper we refer to the above network as the camera-regression network, and refer to our network as the camera-hybrid network. We train both the networks on the augmented training dataset of $\sim108K$ image pairs. Table \ref{tab:camera-loc-val-results} shows the performance of the networks on the validation and the test datasets. We observe that our hybrid architecture performs better than the regression only architecture. Additionally, our network generalizes well to unseen image pairs, and thus performs better on the test dataset.

\subsection{Integrating with Visual Odometry}

We test the performance of integrating visual odometry with our cross-view geolocalization architecture as detailed in section \ref{subsec:integrate-VO}. We use the following as baselines for comparison with our method: visual odometry only; visual odometry integrated with scene localization network; and visual odometry integrated with camera-regression network. For integrating with the scene localization network, we directly use the $k$ satellite image positions instead of the $k$ camera poses for calculating the cross-view geolocalization output. For integrating with the camera-regression network, we obtain the $k$ camera poses from the camera-regression network in instead of our camera-hybrid network.

\begin{figure}[t!]
    \centering
    \includegraphics[width=0.7\linewidth]{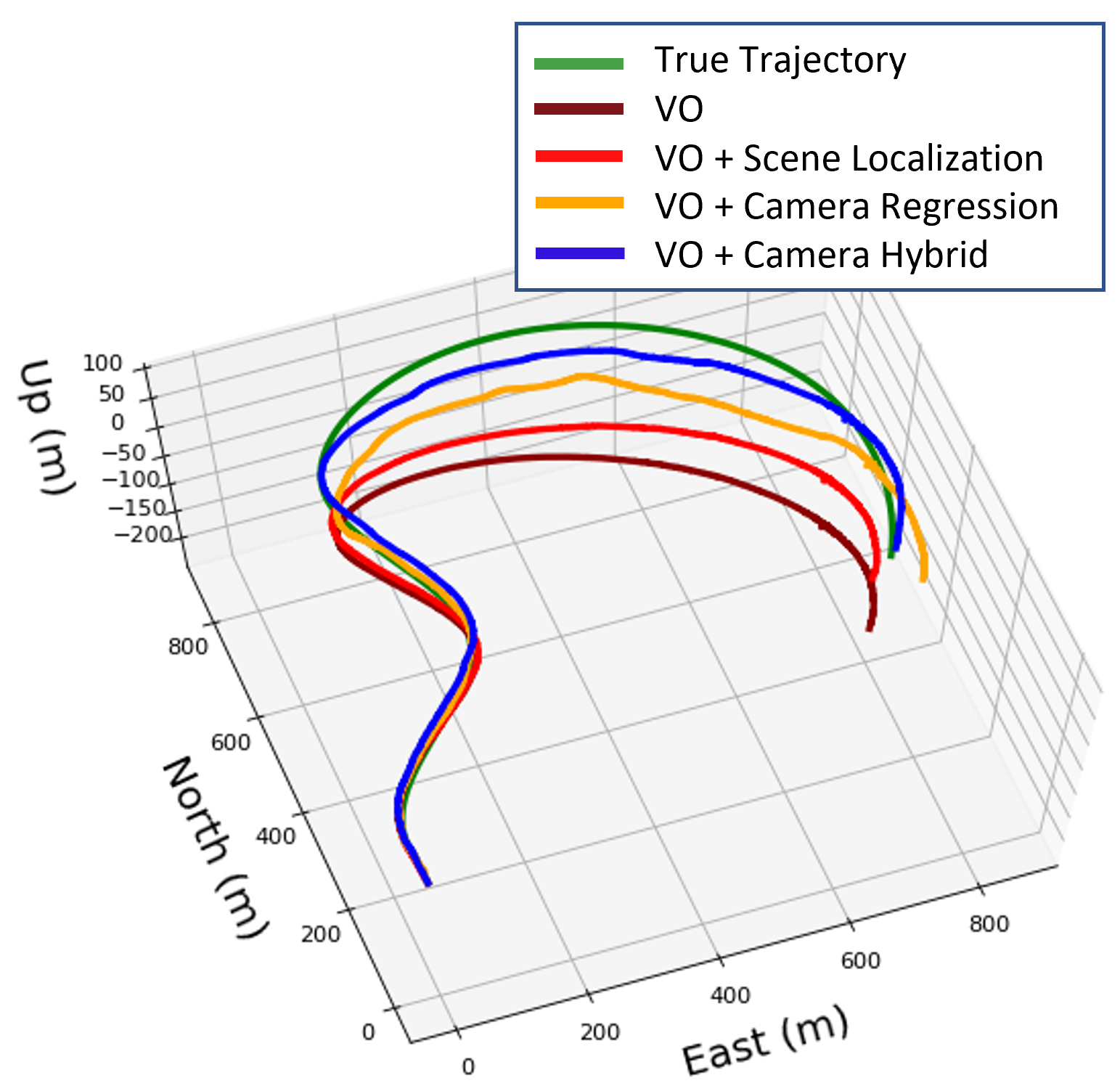}
    \caption{Trajectory estimation results for simulated flight in unseen Detroit dataset. Our cross-view geolocalization combined with visual odometry tracks the true trajectory more accurately compared to other methods.}
    \label{fig:detroit-trajectory-results}
\end{figure}

While integrating our cross-view geolocalization output, we set $k = 9$, ie. for a query UAV image we select $9$ satellite images near the predicted pose. We apply corrections in the KF at a rate of $1$Hz, while the prediction step is performed at $20$Hz. On our configuration of a single node GTX1050 GPU, it takes about $0.45s$ to obtain the output for $9$ image pairs from the scene localization network, and $0.27s$ to obtain the output from our camera-hybrid network. Both the networks need to process the image pairs in parallel, thus making the method achievable in real-time.

Figure \ref{fig:detroit-trajectory-results} shows the trajectory estimation results for the simulated test flight in Detroit. Table \ref{tab:detroit-results} shows the root mean squared errors (RMSE) for the corresponding trajectories. We observe that integrating visual odometry with our camera-hybrid network performs better than the other estimated trajectories. Including global pose corrections from any of the networks improves the trajectory estimation compared to using only visual odometry. However, the corrections from the scene localization network provide a very coarse estimate of the camera position as it assumes the camera to be at the center of the corresponding satellite image. The camera regression and our camera hybrid networks improve on these corrections by additionally estimating the camera transformation from the satellite image to the UAV image.

The orientation errors show smaller changes compared to the position errors. Integrating with the camera-regression and camera-hybrid networks slightly increases the heading errors, since the networks do not perform very well in estimating the heading as seen in Table \ref{tab:camera-loc-val-results}. Table \ref{tab:camera-loc-val-results} also shows lower tilt error range for the networks compared to visual odometry, and hence we observe a slight improvement in the tilt errors.

\begin{table}
\centering
 \begin{tabular}{| C{2.4cm} | C{1.4cm} | C{1.4cm} | C{1.4cm}|} 
 \hline
   & $\mathbf{p}$ RMSE & $\psi$ RMSE & $\theta$ RMSE \\ [0.5ex] 
 \hline
 VO & $141.8m$ $(5.5\%)$ & $3.2^0$ & $10.9^0$ \\
 \hline
 VO + Scene Localization & $97.6m$ $(3.8\%)$ & $3.2^0$ & $10.9^0$ \\
 \hline
 VO + Camera Regression \cite{melekhov2017relative} & $73.3m$ $(2.8\%)$ & $4.6^0$ & $7.5^0$ \\
 \hline
 \textbf{VO + Camera Hybrid} & $\mathbf{36.0m}$ $(\mathbf{1.4\%})$ & $\mathbf{4.0^0}$ & $\mathbf{8.1^0}$ \\
 \hline
\end{tabular}
\caption{Trajectory estimation results for the simulated flight: RMSE in position ($\%$ of trajectory length), heading and tilt. Integrating visual odometry with our method reduces position RMSE significantly.}
\label{tab:detroit-results}
\vspace{0.2cm}
\end{table}

%% file: conclusion.tex
We described our method of cross-view geolocalization to estimate the global pose of a UAV. Our scene localization network successfully learned to map feature vectors based on the contrastive loss. Our hybrid camera localization network performed better than a regression-only network, and also generalized well to unseen test images. While integrating the cross-view geolocalization output with visual odometry, we showed that including a camera localization network in addition to a scene localization network significantly reduced  trajectory estimation errors. One of the limitations of our network was the relatively poor performance in estimating orientation. This paper presented an initial framework for estimating the global camera pose and integrating it with visual odometry. Future work includes testing out different network structures to reduce pose estimation errors, higher number of horizontal cells for camera localization, and expanding the filter to include additional sensors such as inertial units.